\documentclass[journal]{IEEEtran}
\usepackage{xcolor}
\usepackage{gensymb}
\usepackage{amsmath}
\usepackage{xspace}
\usepackage{amsfonts} 
\usepackage{url}
\usepackage{graphicx}
\graphicspath{{figures/}}
\usepackage{lineno}
\usepackage{pifont}
\usepackage{multirow}
\newcommand{\cmark}{\ding{51}}%
\newcommand{\xmark}{\ding{55}}
\newcommand{\textBF}[1]{%
    \pdfliteral direct {2 Tr 0.3 w} %the second factor is the boldness
     #1%
    \pdfliteral direct {0 Tr 0 w}%
}

\begin{document}

\title{A Novel Multi-scale Attention Feature Extraction Block for Aerial Remote Sensing Image Classification}

\author{Chiranjibi Sitaula*+,~\IEEEmembership{Member,~IEEE,}
        Jagannath Aryal+,~\IEEEmembership{Member,~IEEE,}
        Avik Bhattacharya,~\IEEEmembership{Senior Member,~IEEE}
        % <-this % stops a space
\thanks{C Sitaula (*Corresponding) and J Aryal are with the Earth Observation and AI Research Group, Department of Infrastructure Engineering, The University of Melbourne, Parkville, VIC 3010, Melbourne, Australia.  + Equal contribution.}
\thanks{A Bhattacharya is with the Centre of Studies in Resources Engineering (CSRE), Indian Institute of Technology Bombay, Powai, Mumbai-400076, India.}
}
%\linenumbers
%\markboth{IEEE GEOSCIENCE AND REMOTE SENSING LETTERS}
%\newline

\maketitle

\begin{abstract}

Classification of very high-resolution (VHR) aerial remote sensing (RS) images is a well-established research area in the remote sensing community as it provides valuable spatial information for decision-making. 
Existing works on VHR aerial RS image classification produce an excellent classification performance; nevertheless, they have a limited capability to well-represent VHR RS images having complex and small objects, thereby leading to performance instability. 
As such, we propose a novel plug-and-play multi-scale attention feature extraction block (MSAFEB) based on multi-scale convolution at two levels with skip connection, producing discriminative/salient information at a deeper/finer level.  
The experimental study on two benchmark VHR aerial RS image datasets (AID and NWPU) demonstrates that our proposal achieves a stable/consistent performance (minimum standard deviation of $0.002$) and competent overall classification performance (AID: 95.85\% and NWPU: 94.09\%). 

\end{abstract}


\section{Introduction}

%%%%%%%%%% HRV not only outperforms but also has lower complexity than SPP [IEEE ACCESS, GRSL paper] #########################

\IEEEPARstart{V}{ery} high-resolution (VHR) aerial remote sensing (RS) image classification has been increasingly popular  
in the fields of robotics and earth observation owing to the invaluable spatial information they produce. 
Despite significant progress in VHR aerial RS image analytics, such as image retrieval, scene classification, and semantic segmentation, several challenges still exist. These challenges include ambiguous spatial structures, large semantic categories, similar spatial patterns between categories and variable spatial patterns within categories, which negatively impact the spatially explicit classification performance.

Currently, deep learning (DL)-based techniques are dominating the field of VHR aerial RS image classification. These methods have achieved excellent results in several domains, including indoor/outdoor image classification as evidenced by recent study \cite{sitaula2023enhanced}. In these domains, transfer learning has been the focus of DL-based techniques, which are broadly categorised into two processes: one-step \cite{wang2022} (also called end-to-end) and two-step \cite{Weng2017Land-UseFeatures} processes. The one-step process involves a fine-tuning method that utilises either a unimodal-based approach \cite{wang2022} or a multi-modal-based approach (ensemble) \cite{Scott2018EnhancedSets}.
Numerous studies in the literature have utilised the one-step process for VHR aerial RS image classification. For instance, authors in \cite{wang2018scene} utilised the attention recurrent convolutional network (ARCNet) to choose important spatial regions for VHR RS image classification sequentially. Additionally, He et al. \cite{he2019skip} created a skip-connected covariance (SCCov) network for the classification of the VHR aerial RS images and enhanced classification performance by utilising second-order information from the SCCov network. Further, Li et al. \cite{9178726} developed a DL-based approach, called object relationship reasoning convolutional neural network (ORRCNN), to prioritise object relationship importance. 
In another study, Wang et al. \cite{wang2022} established a multi-level feature fusion (MLFF) approach that improved accuracy significantly with the help of a novel adaptive channel dimension reduction approach.
A residual network using a multi-scale feature extraction technique was established by Li et al. \cite{Li2022} that captured the higher-order information, thereby improving the overall classification performance.
Likewise, Wang et al. \cite{9298485} introduced the global-local two-stream architecture, utilising the effectiveness of multi-scale information for image classification. 
In a recent work by Sitaula et al. \cite{sitaula2023enhanced}, an enhanced VHR attention module (EAM) was devised based on the multi-scale information utilising novel enhanced VHR attention mechanism, atrous spatial pyramid pooling (ASPP) \cite{chen2017deeplab} and global average pooling (GAP), which improved the classification performance and robustness.
{
Further, the hybrid approach (Transformer+CNN) \cite{zhao2023localtransformer} that potentially produces billions of parameters were devised for aerial remote sensing scene classification. 
}
The one-step-based approach, which limits extracting salient information at a deeper level, produces excellent classification performance.
Yet it is unstable and non-generalisable, thus requiring to use of a multi-scale convolution at a deeper level with salient information.
%as suggested by the previous study \cite{aryal2023multi}.

While the two-step process has been less studied compared to the one-step process, some studies have shown its effectiveness. For example, Weng et al. \cite{Weng2017Land-UseFeatures} used a pre-trained AlexNet-like architecture on {\it ImageNet} to extract deep features, which were then classified with the help of an extreme learning machine (ELM) algorithm. Similarly, Yu \& Liu \cite{Yu2018AClassification} utilised the VGG-16 and GoogleNet for the feature extraction, where the ELM algorithm was used for the classification. In another study, He et al. \cite{he2018remote} employed pre-trained DL models such as AlexNet and VGG-16 model to extract the deep features, which were stacked and then computed the covariance matrix to obtain correlative information. Then, the support vector machine (SVM) algorithm was used to classify them. Furthermore, Sun et al. \cite{sun2021multi} devised a method to represent images by combining DL-based features with hand-crafted features. These features were then transformed using widely-recognised techniques like scale-invariant scale transform and bag-of-visual-words before being classified using the SVM algorithm. 
{Recently, Xu et al. \cite{9850375} developed hierarchical features fusion of convolutional neural network (HFFCNN), leveraging the multi-scale deep features and producing a bag of visual words representation for the classification.}
{
Further, few-shot learning-based approaches \cite{ma2023multizeroshot,geng2023foregroundzeroshot} were developed to work with limited datasets for the classification.}
%. Those features are encoded with the bag of visual words approach and classified using a linear classifier.}
While the two-step process-based approach imparts excellent results, it not only demands the abundant skills and expertise of multiple machine learning algorithms to attain the maximum benefits but also suffers from performance instability.

Although DL-based techniques provide an excellent classification performance, they have two major limitations: i) performance instability and ii) insufficient representation for VHR RS images having complex and small objects.
Given such limitations, we propose a novel plug-and-play multi-scale attention feature extraction block (MSAFEB) using two-level multi-scale convolution, skip connection, and attention mechanism with the DenseNet201 DL model for the VHR aerial RS image classification.
{Here, the MSAFEB produces deeper multi-scale salient regions (e.g., structures, curves, lines, etc.) of smaller and complex objects, thereby improving separability during classification.}
In summary, the {\bf contributions} of this letter are three-fold:

\begin{enumerate}
    \item [1)] Develop the novel MSAFEB to capture rich discriminative information using a {multi-scale convolution} at two levels with skip connection and the attention mechanism;
    \item [2)] {Evaluate} the MSAFEB {with} eight state-of-the-art (SOTA) pre-trained DL models and show its efficacy through an ablation study;
        {
    \item [3)] Perform the qualitative analysis and explainability of the MSAFEB using Grad-CAM \cite{selvaraju2017grad} and} 
    \item[4)] Demonstrate the superiority of the proposed approach through a comparative study with the SOTA methods {, including multi-scale-based (MS) techniques} on two benchmark datasets.
\end{enumerate}

\section{Proposed Approach}

\begin{figure}
  \centering
  \includegraphics[height=150mm,width=0.50\textwidth, keepaspectratio]{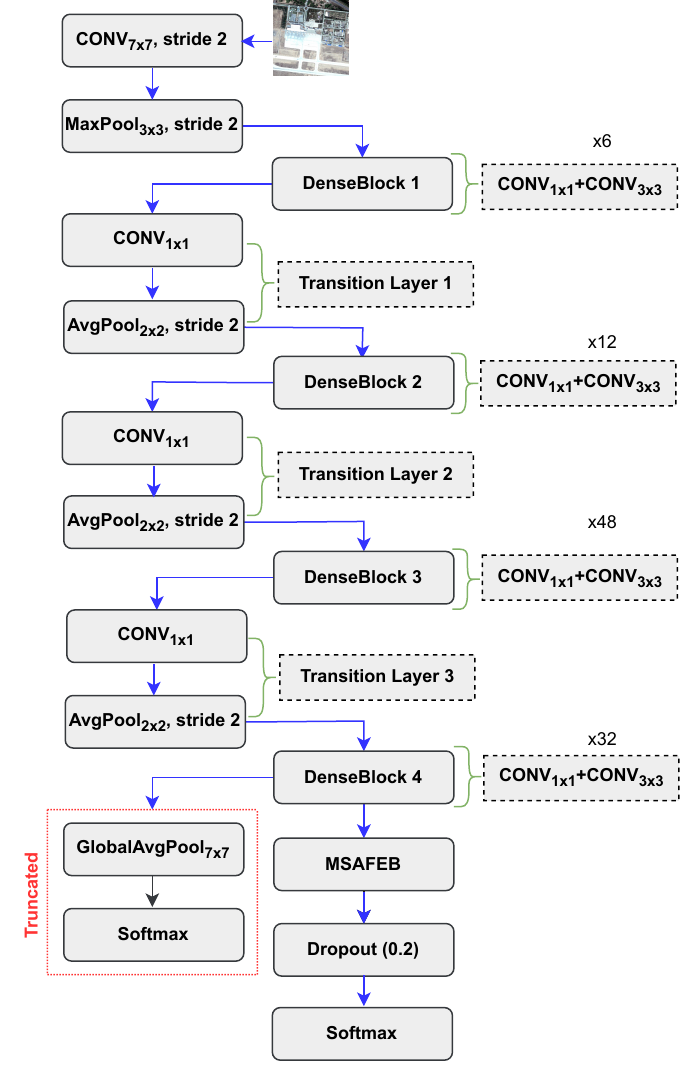}
  \caption{Proposed approach (dashed box: module name/convolutions and x: occurrence of convolutions). 
  %Note that the dashed rectangle denotes the module/layer name or corresponding operation and the 'x' indicates the corresponding operation's occurrence.
  }
  \label{fig:high_level}
\end{figure}

Our proposed approach consists of six steps: 
Sec. \ref{step_1} Feature extraction, Sec. \ref{step_2} Multi-scale convolution, Sec. \ref{step_3} ASPP and GAP, Sec. \ref{step_4} Fusion, Sec. \ref{step_5} Batch normalisation and GAP, and Sec. \ref{step_6} Final feature extraction. Fig. \ref{fig:high_level} shows the overall proposed approach using the DenseNet201 model.

\subsection{Feature extraction}
\label{step_1}

For the feature extraction, we leverage the last layer of DenseBlock 4 of the DenseNet201 model as the input tensor ($I \in \mathbb{R}^{K\times H \times W}$), which has a size of $7 \times 7 \times 1920$ for depth ($K$), height ($H$) and width ($W$), respectively. Here, we resort to the DenseNet201 model as the backbone as it is a less-studied architecture yet provides superior performance than other pre-trained DL models as demonstrated through our ablation study (Sec. \ref{ablation_pre_trained} ).

%EAM
\begin{figure}
  \centering
  \includegraphics[height=60mm,width=0.50\textwidth, keepaspectratio]{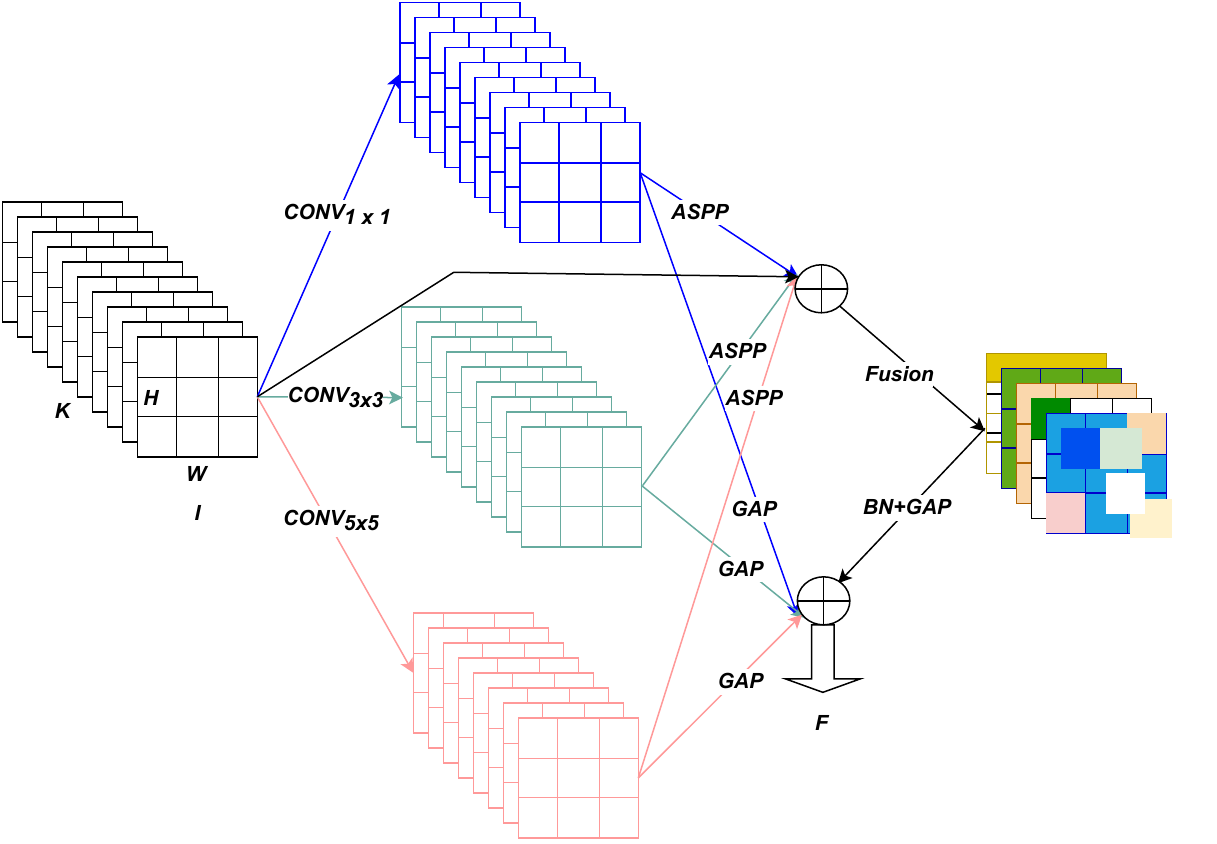}
  \caption{Proposed MSAFEB ($\oplus$: concatenation operator). }
  \label{fig:MSFEB}
\end{figure}

\subsection{Multi-scale convolution}
\label{step_2}

The input tensor ($I$) is processed to achieve multi-scale information as shown in  Fig. \ref{fig:MSFEB}, which acts as the foundation of the proposed MSAFEB. For this, we perform multi-scale convolution ($CONV_{i\times i}(.)$) using three different kernel sizes (Eq. \eqref{eq_combined_conv}). Employing such kernels helps capture the semantic higher-order information from the lower to the higher scale. Note that we empirically set filter size, dilation rate, and groups for each convolution to 480, 4, and 8, respectively. To decrease the training parameters, we also divide the units of each {convolution} by four empirically {without compromising classification performance}.

%%combined equations to save the space
\begin{equation}
    C_i=CONV_{i \times i}(I),
    \label{eq_combined_conv}
\end{equation}
where i $\in$ \{1, 3, 5\} and {$C_i$ denotes the $i^{th}$ convolutional layer with the $i \times i$ kernel size.}

\subsection{ASPP and GAP}
\label{step_3}

Given the effectiveness of multi-scale information at a finer level,
we further utilise dilation {for larger receptive fields that increase the discriminable regions} with multi-scale kernels for three convolution layers {($C_1$, $C_2$ and $C_3$)}. For this, we leverage the $ASPP(.)$ operation \cite{chen2018encoder} that employs four different dilation rates (1, 6, 12, and 18) with two kernel sizes (1 and 3) (Eq. \eqref{eq_combined_aspp}). 

\begin{table}[b]
\scriptsize
\renewcommand{\arraystretch}{1.3}
\caption{Comparative study using overall classification accuracy (OA) (\%) $\,\pm\,$ standard deviation (SD) on AID dataset (Note: Tr/Te=Training/Test; Boldface: best performance and underline: second-best performance% {and MS: multi-scale feature}
). 
%Underlined numbers denote the second-best. 
}
% \label{table_example}
\centering
% \begin{tabular}{|p{2.5cm}|p{2.5cm}|p{2.5cm}|}
\begin{tabular}{p{2.1cm} p{0.2cm} p{2cm} p{2cm}}% p{0.7cm}}
    %\hline
    \hline
    \textBF{Method} &\textBF{{MS?}} &
    \textBF{Tr/Te=50/50}& 
    \textBF{Tr/Te=20/80}\\
    %\textBF{\#Params}
    %\\
    \hline
    % \centering
    % MCNN \cite{} & $91.80\,\pm\,0.22$ & -&- \\
    % TEX-Net-LF \cite{} & $92.96\,\pm\,0.18$ & $90.87\,\pm\,0.11$&- \\
    ARCNet-VGG16 \cite{wang2018scene}&{\xmark} & $93.10\,\pm\,0.550 $ & $88.75\,\pm\,0.400$\\%&17M\\
    % Two-Stream Fusion \cite{} & $94.58\,\pm\,0.25$ & $92.32\,\pm\,0.41$&- \\
    MSCP \cite{he2018remote}&{\xmark} & $94.42\,\pm\,0.170$ & $91.52\,\pm\,0.210$\\%&- \\
    CNN+ELM \cite{Yu2018AClassification}&{\xmark} &$94.58 \pm\, 0.250$ &$92.32\, \pm\, 0.410$\\%^&- \\
    % ResNet\_LGFFE \cite{} & $94.46\,\pm\,0.48$ & $90.83\,\pm\,0.55$&- \\
    LCPB \cite{sun2021multi}&{\xmark} & $91.33\,\pm\,0.360$ & $87.68\,\pm\,0.250$\\%&- \\
    LCPP \cite{sun2021multi}&{\xmark} & $93.12\,\pm\,0.280$ & $90.96\,\pm\,0.330$\\%&- \\
    ORRCNN \cite{9178726}&{\xmark} & $92.00$ & $86.42$\\%- \\
    ACR-MLFF \cite{wang2022}&{\xmark} & $95.06\,\pm\,0.330$ & $92.73\,\pm\,0.120$\\%&35M \\
    MRHNet-50 \cite{Li2022}&{\cmark} & $95.06$ &91.14\\% &- \\
    EAM \cite{sitaula2023enhanced}&{\cmark} & {$95.39\,\pm\,$}\textBF{$0.001$} & 
    {$93.14\,\pm\,0.003$}\\%&- \\
    {HFFCNN \cite{9850375}}&{\cmark} & {$95.32\,\pm\,${$0.240$}} & 
    {$93.08\,\pm\,0.330$}\\%&- \\
    {\bf MSAFEB}&{\cmark} & $\textBF{95.85}\,\pm\,\underline{0.003}$& 
    $\textBF{94.24\,\pm\,{0.002}}$\\   
    %\hline
    \hline
\end{tabular}
\label{tab:AID}
\end{table}

\begin{table}[b] 
\scriptsize
\renewcommand{\arraystretch}{1.3}
\caption{Comparative study using OA (\%) $\,\pm\,$ SD on NWPU dataset ('-': unavailable). 
%The '-' denotes the missing performance measure for the corresponding method/setting.
}
% \label{table_example}
\centering
\begin{tabular}{p{1.8cm} p{0.2cm} p{2cm} p{2cm}}%p{0.7cm}}
    %\hline
    \hline
    \textBF{Method} &\textBF{{MS?}} &
    \textBF{Tr/Te=10/90}& 
    \textBF{Tr/Te=20/80}\\
    %\textBF{\#Params}\\
    \hline
    % \centering
    MSCP \cite{he2018remote}&{\xmark}& $88.07\,\pm\,0.180$ & $90.81\,\pm\,0.130$\\%&- \\
    % SFCNN \cite{} & $89.89\,\pm\,0.16$ & $92.55\,\pm\,0.14$&- \\
    % SDAResNet \cite{} & $89.40$ & $91.15$&- \\
    % VGG\_VD16+SAFF \cite{} & $84.38\,\pm\,0.19$ & $87.86\,\pm\,0.14$&- \\
    SCCov \cite{he2019skip}&{\xmark} & $89.30\,\pm\,0.350$ & $92.10\,\pm\,0.250$\\%&13M \\
    SKAL \cite{9298485}&{\cmark} & {$90.41$}$\,\pm\,0.120$ & $92.95\,\pm\,0.090$\\%&13M \\
    MRHNet-101 \cite{Li2022}&{\cmark} & - & 91.64\\%&- \\
    ACR-MLFF  \cite{wang2022}&{\xmark} & $90.01\,\pm\,0.330$ & $92.45\,\pm\,0.200$\\%&35M \\
    % MRHNet-101 \cite{Li2022} & - &91.64&- \\
    {EAM \cite{sitaula2023enhanced}}&{\cmark} & $90.38\,\pm\,\textBF{0.001}$ &$93.04\,\pm\,\textBF{0.001}$\\%&- \\
    {HFFCNN \cite{9850375}}&{\cmark} & {$87.01\,\pm\,${$0.190$}} & 
    {$90.14\,\pm\,0.200$}\\%&- \\
    {\bf MSAFEB}&{\cmark} & $\textBF{91.71}\,\pm\,\underline{0.002}$& $\textBF{94.09}\,\pm\,\underline{0.002}$\\
    %\hline
    \hline
\end{tabular}
\label{tab:NWPU}
\end{table}
%%%%%%%%%%%%%%%%%%%%%%%Statistical analysis %%%%%%%%%%%%%%%%%%%%%%%%%%

\begin{table*}%[b]
\scriptsize
\renewcommand{\arraystretch}{1.3}
\caption{{Statistical analysis (p-value) of the MSAFEB against SOTA methods using Welch's two-sample t-test method.
% Note that A: ARCNet-VGG16, B: MSCP, C: CNN+ELM. D: LCPB, E: LCPP, F: ORRCNN, G: ACR-MLFF, H: MRHNet-50, I:  EAM, J: HFFCNN, K: SCCov and L: SKAL
%Note that A: ARCNet-VGG16, B: MSCP, C: CNN+ELM. D: LCPB, E: LCPP, G: ACR-MLFF, I:  EAM, J: HFFCNN, K: SCCov and L: SKAL
}} 
%Underlined numbers denote the second-best. 
% \label{table_example}
\centering
%\color{red}
% \begin{tabular}{|p{2.5cm}|p{2.5cm}|p{2.5cm}|}
\begin{tabular}{p{0.6cm} p{0.6cm} p{1.5cm} p{0.9cm}p{1cm} p{0.8cm} p{0.8cm} p{1.2cm}p{0.8cm} p{1.3cm}p{0.9cm} p{0.9cm}}% p{0.7cm}}
    %\hline
    \hline
   Dataset&
   Setting &
   ACRNet-VGG16 \cite{wang2018scene}&
   MSCP \cite{he2018remote} &
   CNN+ELM \cite{Yu2018AClassification}&
   LCPB \cite{sun2021multi}&
   LCPP \cite{sun2021multi}&
   %F \cite{9178726} &
   ACR-MLFF \cite{wang2022}&
   %H \cite{Li2022}&
   EAM \cite{sitaula2023enhanced}&
   HFCNN \cite{9850375} &
   SCCov \cite{he2019skip}&
   SKAL \cite{9298485} \\%&
  % M \cite{Li2022}\\
  \hline
  %\centering
    \multirow{2}{*}{AID} & 50/50 &7.1e-08&7.1e-10&6.1e-08&1.3e-10&1.9e-10&3.4e-05&5.8e-11&6.4e-05& -&-%&- 
    \\
     & 20/80 &9.1e-12& 1.5e-11&1.2e-07&2.6e-14& 1.6e-10&1.9e-11&2.2e-16&1.4e-06& -&-%&- 
     \\
     \hline
      \multirow{2}{*}{NWPU} & 10/90 &-& 2.7e-13&-&-& -&5.4e-08&2.9e-15&4.5e-14& 4.2e-09&7.4e-11%&- 
      \\
     & 20/80 &-&3.7e-14&-& -&-&9.0e-10&9.7e-15&3.4e-13& 1.1e-09&1.8e-11%&- 
     \\
    \hline
    %\hline
\end{tabular}
\label{tab:statistical_analysis}
\end{table*}

%%%%%%%%%%%%%%%%%%%%%%ablative study%%%%%%%%%%%%%%%%%%%%%%%%%%%%%%%%%%%
%%%%%%%%%%%%%%%%%%%%%%%%%%%%%%%%%%%%%%%%%%%%%%%%%%%%%%%%%%%%%%%%%%%%%%%%
\begin{table*}[tb]%[!t]
\renewcommand{\arraystretch}{1.3}
\scriptsize
\caption{Ablation study of MSFEB [with (w/) and without(w/o)] on eight SOTA pre-trained DL models (A1: VGG-16, A2: VGG-19, A3: ResNet-50, A4: Inception-V3, A5: InceptionRes-V2, A6: MobileNet-V2, A7: EfficientNetB0, A8: DenseNet201) using OA (\%) $\,\pm\,$ SD.}
% \label{table_example}
\centering
\begin{tabular}{p{0.6cm} p{1.7cm} p{1.7cm}|p{1.7cm} p{1.7cm}|p{1.7cm} p{1.7cm}|p{1.7cm} p{1.7cm}}
   % \hline
    \hline
    & \multicolumn{4}{c|}{AID} & \multicolumn{4}{c}{NWPU}\\
     \cline{2-9}
      \textBF{Model}&
      \multicolumn{2}{c|}{ \textBF{Tr/Te=20/80 }}&
      \multicolumn{2}{c|}{\textBF{Tr/Te=50/50 }}&
       \multicolumn{2}{c|}{ \textBF{Tr/Te=10/90 }}&
      \multicolumn{2}{c}{\textBF{Tr/Te=20/80 }} \\
      \cline{2-9}
      %\cline{4-5}
      % \multicolumn{2}{c}{C (\%)} \\
    &\textBF{w/}& 
    \textBF{w/o}&
    \textBF{w/} & 
    \textBF{w/o}&
    \textBF{w/}& 
    \textBF{w/o}&
    \textBF{w/} & 
    \textBF{w/o}    
    \\
    % \cline{2-5}
    \hline
    %EAM and ASPP
    A1&$\textBF{87.64}\,\pm\,\textBF{0.008}$&$84.32\,\pm\,0.010$&$\textBF{92.95}\,\pm\,{0.008}$&$91.42\,\pm\,\textBF{0.007}$&$\textBF{80.74}\,\pm\,\textBF{ 0.020}$&$28.86\,\pm\,0.326$&$\textBF{88.56}\,\pm\,\textBF{ 0.004}$&$18.94\,\pm\,0.330$\\ 
    A2&$\textBF{82.95}\,\pm\,\textBF{0.030}$&$46.41\,\pm\,{0.340}$&$\textBF{90.33}\,\pm\,\textBF{ 0.017}$&$86.72\,\pm\,0.050$&$\textBF{74.02}\,\pm\,\textBF{ 0.055}$&$02.23\,\pm\,\textBF{0.001}$&$\textBF{81.21}\,\pm\,\textBF{ 0.035}$&$14.91\,\pm\,{0.250}$\\ 
    A3&$\textBF{92.76}\,\pm\,\textBF{0.001}$&$92.66\,\pm\,0.002$&$\textBF{95.57}\,\pm\,{0.003}$&${95.16}\,\pm\,\textBF{0.002}$&$\textBF{90.36}\,\pm\,\textBF{ 0.002}$&${89.70}\,\pm\,{0.003}$&$\textBF{93.04}\,\pm\,\textBF{ 0.001}$&${92.78}\,\pm\,\textBF{0.001}$\\
    A4&$91.42\,\pm\,0.002$&$\textBF{91.58}\,\pm\,\textBF{ 0.001}$&${94.64}\,\pm\,{0.003}$&$\textBF{95.60}\,\pm\,\textBF{ 0.001}$&${88.10}\,\pm\,\textBF{0.002}$&$\textBF{88.24}\,\pm\,\textBF{ 0.002}$&${91.63}\,\pm\,\textBF{0.001}$&$\textBF{91.79}\,\pm\,\textBF{ 0.001}$\\
    A5&${90.36}\,\pm\,\textBF{0.005}$&$\textBF{90.62}\,\pm\,\textBF{ 0.005}$&$\textBF{93.55}\,\pm\,\textBF{0.004}$&${93.53}\,\pm\,{0.005}$&${86.88}\,\pm\,\textBF{0.004}$&$\textBF{87.82}\,\pm\,\textBF{ 0.004}$&$\textBF{90.59}\,\pm\,{0.006}$&${90.36}\,\pm\,\textBF{0.005}$\\ 
    A6&$\textBF{89.66}\,\pm\,\textBF{0.004}$&$\textBF{89.66}\,\pm\,\textBF{0.004}$&$\textBF{92.85}\,\pm\,\textBF{0.004}$&{$92.69\,\pm\,0.006$}&$\textBF{86.50}\,\pm\,\textBF{0.002}$&{$86.16\,\pm\,0.002$}&$\textBF{90.58}\,\pm\,\textBF{0.000}$&$90.25\,\pm\,\textBF{0.000}$\\ 
    A7&$\textBF{91.34}\,\pm\,{0.002}$&${90.08}\,\pm\,\textBF{0.001}$&${94.79}\,\pm\,\textBF{0.001}$&$\textBF{95.01}\,\pm\,{0.004}$&$\textBF{87.45}\,\pm\,{\bf 0.001}$&${86.91}\,\pm\,{0.002}$&$\textBF{91.70}\,\pm\,{0.003}$&${91.51}\,\pm\,\textBF{0.001}$\\
    A8&$\textBF{94.24}\,\pm\,{0.002}$&${93.97}\,\pm\,\textBF{0.001}$&${95.85}\,\pm\,{0.003}$&$\textBF{96.08}\,\pm\,{\bf 0.001}$&$\textBF{91.71}\,\pm\,{\bf 0.002}$&${91.64}\,\pm\,\textBF{0.002}$&$\textBF{94.09}\,\pm\,\textBF{0.001}$&${94.07}\,\pm\,\textBF{0.001}$\\
   \hline
%    \hline
\end{tabular}
\label{tab: ablative}
\end{table*}

%%%%%combined equations
{
\begin{equation}
    D_i=ASPP(C_i),
    \label{eq_combined_aspp}
\end{equation}}
where $D_i$ denotes the resultant feature map after $ASPP(.)$ operation on $C_i$. 
% %%%%%combined equations
% \begin{equation}
%     D_i=ASPP(C_i),
%     \label{eq_combined_aspp}
% \end{equation}
% where $D_i$ denotes the resultant feature map after $ASPP(.)$ operation on $C_i$. Meanwhile, we perform the $GAP(.)$ operation over three convolution layers (Eq. \eqref{eq:combined_gap}).
% %over $C_1$, $C_2$, and $C_3$, respectively.).
{Then, we perform the GAP operations for the feature aggregation over the convolutional features ($C_i$) (Eq. \eqref{eq:combined_gap}.
%, consisting of single-level multi-scale information only as complementary information.
}
%%%%%combined
\begin{equation}
    G_i=GAP(C_i),
    \label{eq:combined_gap}
\end{equation}
{where $G_i$ denotes the resultant GAP-based features of the $C_i$.}

%\subsection{$CONV_{1 \times 1}$ and EAM}
\subsection{Fusion}
\label{step_4}

The multi-scale features obtained from the previous {steps} are combined to achieve unified multiple information for the input image. For this, we use a simple $1 \times 1$ convolution operation ($CONV_{1\times1}(.)$) over the concatenated tensors. For the concatenation ($\oplus$), we use three ASPP-based tensors ($D_1$, $D_2$, $D_3$) and input tensor ($I$) as a skip-connection. After the concatenation, we apply the $EAM(.)$ proposed by Sitaula et al. \cite{sitaula2023enhanced} on top of such concatenated tensors (Eq. \eqref{eq:7}). The $EAM(.)$ operation helps capture the highly semantic regions {over resultant tensor} using its enhanced convolution with attention approach.

{
\begin{equation}    
    E=EAM(CONV_{1\times 1}(I\oplus D_1\oplus D_2\oplus D_3)),
    \label{eq:7}
\end{equation}}

where $E$ denotes the resultant multi-scale tensor with attention (salient) information.

{
\subsection{Batch normalisation and GAP}
\label{step_5}
The attention-based tensor is fed to the batch normalisation layer ($BN(.)$) to accelerate the training process, thereby reducing the convergence time. Then, the $GAP(.)$ operation is applied over the resultant tensor for aggregation (Eq. \eqref{eq:8}).
\begin{equation}    
    G=GAP(BN(E)),
    \label{eq:8}
\end{equation}
where $G$ denotes the multi-scale aggregated tensor.
% and {$G_i$ denotes the GAP-based features of the $i^{th}$ convolutional layers after $ASPP(.)$ operation as shown in Eq. \ref{eq_combined_aspp}
% }.
}

\subsection{Final feature extraction}
\label{step_6}

At last, we concatenate the multi-scale aggregated tensor ($G$) with the GAP-based features obtained from step \ref{step_3} (Eq. \eqref{eq:fusion}.
% convolution with 1x1, 3x3, and 5x5 kernel sizes ($G_1$, $G_2$, and $G_3$) (Eq. \eqref{eq:fusion}).

\begin{equation}
F=({G}\oplus G_1\oplus G_2\oplus G_3),
    \label{eq:fusion}
\end{equation}

where $F$ denotes the final multi-scale attention features to be fed to the {\it Dropout} layer and {\it Softmax} layer for the classification as shown in Fig. \ref{fig:high_level}.

\subsection{Datasets}

We utilise two benchmark VHR aerial RS scene datasets: the Aerial Image Data Set (AID) \cite{xia2017aid} and NWPU-RESISC45 (also called NWPU) \cite{cheng2017remote}. The AID contains 10,000 images, which are divided into 30 categories, with image sizes of $600 \times 600$ pixels and spatial resolutions ranging from 0.5 to 8 m. The NWPU has 31,500 images divided into 45 categories, each with an image size of $256 \times 256$ pixels and spatial resolutions varying from 0.2 to 30 m.

We use the standard train/test splits as in existing work \cite{sitaula2023enhanced} for the evaluation. 
%{For this, we prepare random five folds for each setting on both datasets and report the averaged performance.}
For the AID dataset, we create five random splits each for 50/50 and 20/80 settings. Furthermore, we create five random splits each under 10/90 and 20/80 settings for the NWPU dataset. The overall classification results are reported as the average of these splits for each setting of the dataset.

\subsection{Implementation}

We use the Keras package implemented in Python, running on an NVIDIA GeForce RTX 3070 Ti GPU Laptop. 
For the optimisation, we use the {\it Adam} optimiser, with a step-wise weight decay penalty of {$1 \times 10^{-4}$} and a batch size of {16}. 
We set the learning rate to $3 \times 10^{-3}$ and trained for 50 epochs. We utilise the Early Stopping criteria with a patience of 5 for the validation loss during the model training.
Further, we implement data augmentation techniques using horizontal flipping, random cropping, and shuffling.

\subsection{Results and discussion}
\subsubsection{Comparative study with the SOTA methods}

We conduct a comparative study of our proposed approach with SOTA methods on the AID and NWPU datasets. Tables \ref{tab:AID} and \ref{tab:NWPU} present the experimental results on the AID and NWPU, respectively.
We achieve the highest classification accuracy of 95.85\% (0.46\% higher than the second-best (EAM \cite{sitaula2023enhanced})) and 94.24\% (1.10\% higher than the second-best (EAM \cite{sitaula2023enhanced})) on the AID dataset in Table \ref{tab:AID} for the 50/50 and 20/80 settings, respectively. Further, our approach consistently delivers a comparable standard deviation (second-best) of 0.003 for the 50/50 setting and a lowest standard deviation of 0.002 for the 20/80 setting. This indicates that our approach imparts a stable/consistent performance.
Table \ref{tab:NWPU} further highlights our approach's superiority, with encouraging classification accuracies of 91.71\% for the 10/90 setting (1.30\% higher than the second-best (SKAL \cite{9298485})) and 94.09\% for the 20/80 setting (1.05\% higher than the second-best (EAM \cite{sitaula2023enhanced})). Additionally, the proposed approach imparts the second-best standard deviation of 0.002 on both settings on the NWPU dataset, further underscoring our approach's consistency/stability.
{The MSAFEB produces larger receptive fields accompanied by the attention mechanism for performance improvement; nevertheless, it has been missing in the existing MS-based SOTA.}
{Further, our approach delivers extremely significant performance (p-value$<$0.05) while using student t-test (Welch's two-sample t-test) against SOTA methods (Table \ref{tab:statistical_analysis}). }
%This further underlines the superiority.}

\subsubsection{Ablation study}
\label{ablation_pre_trained}

\begin{figure}
  \centering
  \includegraphics[height=65mm,width=0.48\textwidth, keepaspectratio]{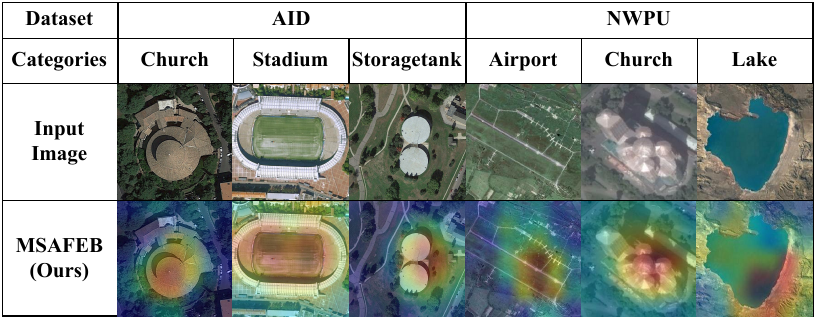}
  \caption{Grad-CAM feature map produced by the MSAFEB. }
  \label{fig:featuremap}
\end{figure}

We analyse the efficacy of the proposed MSAFEB with eight off-the-shelf pre-trained DL models (VGG-16, VGG-19, ResNet-50, Inception-V3, InceptionRes-V2, MobileNet-V2, EfficientNetB0 and DenseNet201) on both datasets. Here, we utilise the last layer of each topless pre-trained DL model, which is enabled and achieved using the 'include\_top=False' setting during implementation. For the experiment, we consider two scenarios: i) with MSAFEB and ii) without MSAFEB.
In the first scenario, we stack the MSAFEB on the last layer of the DL model for the classification, whereas we perform the GAP operation of such layer, which is a widely-accepted approach for transfer learning with pre-trained models, and proceed to the classification in the second scenario. Table \ref{tab: ablative} presents the experimental results for scenarios on both datasets.  These results show that the performance of the DL model without MSAFEB is mostly lower than that with MSAFEB (highest margin of 71.79\% with the VGG-19 and lowest margin of 0.02\% with the InceptionRes-V2 and DenseNet201) as it has the capability to attain highly discriminative information from a deeper level as seen in Fig. \ref{fig:featuremap}.
Note that the second scenario produces a lower performance potentially due to insufficient discriminative information than ours despite having a comparable consistent performance while training the model from the first layer. 
{In contrast, the first scenario (ours) struggles to identify the dissimilar object patterns (e.g., smaller planes with varying structures in the 'Airport' category) present in the same image as shown in Fig. \ref{fig:featuremap}. 
Furthermore, it incurs comparatively larger parameters than the one without MSAFEB (e.g., 34.9 million and 18.1 million parameters for the first and second scenarios, respectively with the DenseNet201).% and follows a similar pattern for other DL models.
}

%% Feature map
\section{Conclusion}

In this letter, we introduce an innovative multi-scale attention feature extraction block (MSAFEB), which has the capability to produce more discriminative multi-scale salient information at a deeper level than previous methods, with the DenseNet201 model to classify VHR RS images. The introduction of the MSAFEB embraced accuracy without compromising consistency.
Our proposed and implemented approach delivers excellent performance on two benchmark datasets (AID: 95.85\% and NWPU: 94.09\%) with consistency (minimum standard deviation of 0.002). 
For future work, we suggest exploring the MSAFEB with several other well-established attention mechanisms (e.g., convolutional block attention module, bottleneck attention module, spatial excitation module), which deliver different kinds of salient information, using VHR RS datasets.
Also, it is envisioned that the proposed approach could be potentially transferred to use cases in precision agriculture such as crop monitoring, drought and stress monitoring, utilising intelligent land use and land cover classification techniques.

\bibliographystyle{ieeetr}
\bibliography{IEEEabrv,references_.bib}
\end{document}